\documentclass[letterpaper]{article} 
\usepackage{aaai18}  
\usepackage{times}  
\usepackage{helvet}  
\usepackage{courier}  
\usepackage{url}  
\usepackage{graphicx}  
\newcommand{\citet}[1]
{\citeauthor{#1}~\shortcite{#1}}
\newcommand{\citep}{\cite}

\usepackage{amsmath} 
\usepackage{amsfonts}
\usepackage{bm}
\usepackage{tikz}
\usepackage[T1]{fontenc}

\def\ci{\perp\!\!\!\perp}
\newcommand{\T}{\theta}

\title{Fair Inference on Outcomes} 
\author{Razieh Nabi \ and Ilya Shpitser\\
Computer Science Department\\
Johns Hopkins University  \\
$(rnabiab1@, ilyas@cs).jhu.edu$
}

\begin{document}
\maketitle

\begin{abstract}

In this paper, we consider the problem of fair statistical inference involving outcome variables.  Examples include classification and regression problems, and estimating treatment effects in randomized trials or observational data.  The issue of fairness arises in such problems where some covariates or treatments are ``sensitive,'' in the sense of having potential of creating discrimination.  In this paper, we argue that the presence of discrimination can be formalized in a sensible way as the presence of an effect of a sensitive covariate on the outcome along certain causal pathways, a view which generalizes \citep{pearl09causality}. A fair outcome model can then be learned by solving a constrained optimization problem. We discuss a number of complications that arise in classical statistical inference due to this view and provide workarounds based on recent work in causal and semi-parametric inference. 

\end{abstract}

\section{Introduction}
\label{sec:intro}

As statistical and machine learning models become an increasingly ubiquitous part of our lives, policymakers, regulators, and advocates have expressed concerns about the impact of deployment of such models that encode potential harmful and discriminatory biases. Unfortunately, data analysis is based on statistical models that do not, by default, encode human intuitions about fairness and bias.  For instance, it is well-known that recidivism is predicted at higher rates among certain minorities in the US \citep{propublica}. To what extent are these predictions discriminatory?  What is a sensible framework for thinking about these issues?  
A growing community is now addressing issues of fairness and transparency in data analysis in part by defining, analyzing, and mitigating harmful effects of algorithmic bias from a variety of perspectives and frameworks \citep{pedreshi08discrimination,feldman15certifying,hardt16equality,kamiran13quantifying,corbett-davies17algorithmic,jabbari16fair}.

In this paper, we propose to model discrimination based on a ``sensitive feature,'' such as race or gender, with respect to an outcome as the presence of an effect of the feature on the outcome along certain ``disallowed'' causal pathways.  As a simple example, discussed in \citep{pearl09causality}, job applicants' gender should not \emph{directly} influence the hiring decision, but may influence the hiring decision indirectly, via secondary applicant characteristics important for the job, and correlated with gender.  We argue that this view captures a number of intuitive properties of discrimination, and generalizes existing formal \citep{pearl09causality,Zhang17causal} and informal proposals \citep{Bertrand2004}.


The paper is organized as follows. We first fix our notation and give a brief introduction to causal inference and mediation analysis, which will be necessary to formally define our approach to fair inference. We then discuss representative prior work on fair inference, and enumerate issues these methods may run into. Moving forward, we show that fair inference from finite samples under our definition can be viewed as a certain type of constrained optimization problem. We then discuss a number of complications to the basic framework of fair inference. We illustrate our framework via experiments on real datasets in the experimental section followed by additional discussion and final conclusions. 

\section{Notation And Preliminaries} 
\label{sec:prelim}

Variables will be denoted by uppercase letters, $V$, values by lowercase letters, $v$, and sets by bold letters. A state space of a variable will be denoted by ${\mathfrak X}_V$. We will represent datasets by ${\cal D} = (Y, {\bf X})$, where $Y$ is the outcome and ${\bf X}$ is the feature vector.
We denote by $x^i_j$ and $y^i$ the $i$th realization of the $j$th feature $X_j \in {\bf X}$ and the outcome $Y$.  Similarly,
${\bf x}^i$ is the $i$th realization of the entire feature vector. 

In this paper, we consider probabilistic classification and regression problems with a set of features ${\bf X}$ and an outcome $Y$, where a feature $S \in {\bf X}$ is \emph{sensitive}, in the sense that making inferences on the outcome $Y$ based on $S$ carelessly may result in discrimination. There are many examples of $S,Y$ pairs that have this property.  These include hiring discrimination ($Y$ is a hiring decision, and $S$ is gender), or recidivism prediction in parole hearings (where $Y$ is a parole decision, and $S$ is race). Our approach readily generalizes to any outcome based inference task, such as establishing causal effects, although we do not consider these generalizations here in the interests of space.

\section{Causal Inference}
\label{sec:causal}

In causal inference, in addition to the outcome $Y$, we distinguish a \emph{treatment variable} $A \in {\bf X}$, and sometimes also one or more \emph{mediator variables} $M \in {\bf X}$, or ${\bf M} \subseteq {\bf X}$.  The primary object of interest in causal inference is the potential outcome variable, $Y(a)$ \citep{neyman23app}, which represents the outcome if, possibly contrary to fact, $A$ were set to value $a$. Given $a,a' \in {\mathfrak X}_A$, comparison of $Y(a')$ and $Y(a)$ in expectation: $\mathbb{E}[Y(a)] - \mathbb{E}[Y(a')]$ would allow us to quantify the average causal effect (ACE) of $A$ on $Y$.  In general, the average causal effect is not computed using the conditional expectation $\mathbb{E}[Y | A]$, since association of $A$ and $Y$ may be spurious or only partly causal. 

Causal inference uses assumptions in \emph{causal models} to link observed data with counterfactual contrasts of interest. When such a functional exists, we say the parameter is \emph{identified} from the observed data under the causal model. One such assumption, known as \emph{consistency}, states that the mechanism that determines the value of the outcome does not distinguish the method by which the treatment was assigned, as long as the treatment value assigned was invariant.  This is expressed as $Y(A) = Y$.  Here $Y(A)$ reads ``the random variable $Y$, had $A$ been intervened on to whatever value $A$ would have naturally attained.''  

Another standard assumption is known as \emph{conditional ignorability}.  This assumption states that conditional on a set of factors ${\bf C} \subseteq {\bf X}$, $A$ is independent of any counterfactual outcome, i.e. $Y(a) \ci A | {\bf C}, \forall a \in {\mathfrak X}_A$, where $(.\ci.|.)$ represents conditional independence. Given these assumptions, we can show that $p(Y(a)) 
= \sum_{\bf C} p(Y \mid a, {\bf C}) p({\bf C})$, 
known as the adjustment formula, the backdoor formula, or stratification.
Intuitively, the set ${\bf C}$ acts as a set of \emph{observed confounders}, such that adjusting for their influence suffices to remove all non-causal dependence of $A$ and $Y$, leaving only the part of the dependence that corresponds to the causal effect. 
A general characterization of identifiable functionals of causal effects exists \citep{tian02onid,shpitser07hierarchy}.

\subsection{Causal Diagrams}
\label{subsec:graphs}

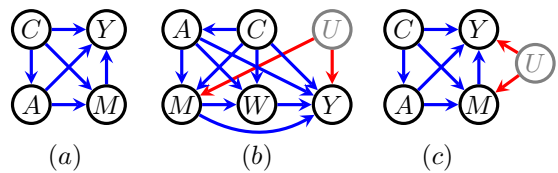
\begin{figure}[t]
\centering
\begin{tikzpicture}[>=stealth, node distance=1.0cm]
    \tikzstyle{format} = [draw, very thick, circle, minimum size=5.0mm,
	inner sep=0pt]
	\begin{scope}
		\path[->, very thick]
			node[format] (A) {$A$}
			node[format, right of= A] (M) {$M$}
			node[format, above of= A] (C) {$C$}
			node[format, above of= M] (Y) {$Y$}
			
			(A) edge[blue] (M)
			(M) edge[blue] (Y)
			(A) edge[blue] (Y)
			(C) edge[blue] (A)
			(C) edge[blue] (Y)
			(C) edge[blue] (M)
			
			node[below of=A, yshift=0.3cm, xshift=0.45cm] (l) {$(a)$}
	   ;
	\end{scope}
	\begin{scope}[xshift=2.0cm]
		\path[->, very thick]
			node[format] (M) {$M$}
			node[format, above of=M] (A) {$A$}
			node[format, right of=M] (W) {$W$}
			node[format, right of=W] (Y) {$Y$}
			node[format, gray, above of=Y] (U) {$U$}
			node[format, above of=W] (C) {$C$}
			
			(M) edge[blue, bend right] (Y)
			(A) edge[blue] (M)
			(M) edge[blue] (W)
			(W) edge[blue] (Y)
			(U) edge[red] (M)
			(U) edge[red] (Y)
			(C) edge[blue] (A)
			(C) edge[blue] (M)
			(C) edge[blue] (W)
			(C) edge[blue] (Y)
			(A) edge[blue] (Y)
			(A) edge[blue] (W)
			
			node[below of=W, yshift=0.3cm, xshift=0.0cm] (l) {$(b)$}
	   ;
	\end{scope}
	\begin{scope}[xshift=4.95cm]
		\path[->, very thick]
			node[format] (A) {$A$}
			node[format, right of= A] (M) {$M$}
			node[format, above of= A] (C) {$C$}
			node[format, above of= M] (Y) {$Y$}
			node[format, gray, right of= Y, xshift=-0.25cm, yshift=-0.45cm] (U) {$U$}
			
			(A) edge[blue] (M)
			(M) edge[blue] (Y)
			(A) edge[blue] (Y)
			(C) edge[blue] (A)
			(C) edge[blue] (Y)
			(C) edge[blue] (M)
			
			(U) edge[red] (M)
			(U) edge[red] (Y)
			
			node[below of=A, yshift=0.3cm, xshift=0.45cm] (l) {$(c)$}
	   ;
	\end{scope}
\end{tikzpicture} 
\caption{(a) A causal graph with a single mediator.
(b) A causal graph with two mediators, one confounded with the outcome via an unobserved common cause.
(c) A causal graph with a single mediator where the natural direct effect is not identified.} 
\label{fig:med}
\end{figure}

Causal relationships are often represented by graphical causal models \citep{spirtes01causation,pearl09causality}.  Such models generalize independence models on directed acyclic graphs, also known as Bayesian networks \citep{pearl88probabilistic}, to also encode conditional independence statements on counterfactual random variables \citep{thomas13swig}.  In such graphs, vertices represent observed random variables, and absence of directed edges represents absence of direct causal relationships. 
As an example, in Fig.~\ref{fig:med} (a), $C$ is potentially a direct cause of $A$, while $M$ mediates a part of the causal influence of $A$ on $Y$, represented by all directed paths from $A$ to $Y$.

\subsection{Mediation Analysis}
\label{sec:mediation}

A natural step in causal inference is understanding the mechanism by which $A$ influences $Y$. A simple form of understanding mechanisms is via mediation analysis, where the causal influence of $A$ on $Y$, as quantified by the ACE, is decomposed into the \emph{direct effect}, and the \emph{indirect effect} mediated by a mediator variable $M$.  
In typical mediation settings, ${\bf X}$ is partitioned into a treatment $A$, a single mediator $M$, an outcome $Y$, and a set of baseline factors ${\bf C} = {\bf X} \setminus \{ A, M, Y \}$.

Mediation is encoded via a counterfactual contrast using a nested potential outcome of the form $Y(a,M(a'))$, for $a,a' \in {\mathfrak X}_A$.  $Y(a,M(a'))$ reads as
``the outcome $Y$ if $A$ were set to $a$, while $M$ were set to whatever value it would have attained had $A$ been set to $a'$.  An intuitive interpretation for this counterfactual occurs in cases where a treatment can be decomposed into two disjoint parts, one of which acts on $Y$ but not $M$, and another acts on $M$ but not $Y$.  For instance, smoking can be decomposed into smoke and nicotine.  Then if $M$ is a mediator affected by smoke, but not nicotine (for instance lung cancer), and $Y$ is a composite health outcome, then $Y(a,M(a'))$ corresponds to the response of $Y$ to an intervention that sets the nicotine exposure (the part of the treatment associated with $Y$) to what it would be in smokers, and the smoke exposure (the part of the treatment associated with $M$) to what it would be in non-smokers. An example of such an intervention would be a nicotine patch.

Given $Y(a,M(a'))$, we define the following effects on the mean difference scale: \emph{the natural direct effect (NDE)} as $\mathbb{E}[Y(a,M(a'))] - \mathbb{E}[Y(a')]$, and \emph{the natural indirect effect (NIE)} as $\mathbb{E}[Y(a)] - \mathbb{E}[Y(a,M(a'))]$ \citep{robins92effects}.  Intuitively, the NDE compares the mean outcome affected only by the part of the treatment that acts on it, and the mean outcome under placebo treatment.  Similarly, the NIE compares the outcome affected by all treatment, and the outcome where the part of the treatment that acts on the mediator is ``turned off.''  A telescoping sum argument implies that ACE $=$ NDE $+$ NIE.

Aside from consistency, additional assumptions are needed to identify $p(Y(a,M(a')))$. One such assumption is known as \emph{sequential ignorability}, and states that conditional on ${\bf C}$, counterfactuals $Y(a,m)$ and $M(a')$ are independent for any $a,a' \in {\mathfrak X}_A$, $m \in {\mathfrak X}_M$. In addition, conditional ignorability for $Y$ acting as the outcome, and $A,M$ acting as a single composite treatment, that is
  $(Y(a,m) \ci A,M \mid {\bf C})$, and conditional ignorability for $M$ acting as the outcome, that is $(M(a') \ci A \mid {\bf C})$, should hold. Under these assumptions, the NDE is identified as the functional known as the mediation formula \citep{pearl11cmf}:
\begin{small}
\begin{align}
\sum_{{\bf C},M} \left( \mathbb{E}[Y | a,M,{\bf C}] - \mathbb{E}[Y | a',M,{\bf C}] \right) p(M \mid a',{\bf C}) p({\bf C}), \label{eqn:edge-g}
\end{align}
\end{small}
which may be estimated by plug in estimators, or other methods \citep{tchetgen12semi}.

\subsection{Path-Specific Effects}
\label{subsec:path}
 
In general, we may be interested in decomposing the ACE into effects along particular causal pathways.  For example in Fig.~\ref{fig:med} (b), we may wish to decompose the effect of $A$ on $Y$ into the contribution of the path $A \to W \to Y$, and the
path bundle $A \to Y$ and $A \to M \to W \to Y$.  Effects along paths, such as an effect along the path $A \to W \to Y$, are known as \emph{path-specific effects} \citep{pearl01direct}.  Just as the NDE and NIE, path-specific effects (PSEs) can be formulated as nested counterfactuals \citep{shpitser13cogsci}. The general idea is that along the pathways of interest, variables behave as if the treatment variable $A$ were set to the ``active value'' $a$, and along other pathways, variables behave as if the treatment variable $A$ were set to the ``baseline value'' $a'$, thus ``turning the treatment off.''  Using this scheme, the path-specific effect of $A$ on $Y$ along the path $A \to W \to Y$, on the mean difference scale, can be formulated as
\begin{align}
\mathbb{E}[Y(a', W(M(a'),a), M(a'))] - \mathbb{E}[Y(a')]
\label{eqn:path-eff}
\end{align}

Under a more complex sets of assumptions found in \citep{shpitser13cogsci}, the counterfactual mean $\mathbb{E}[Y(a', W(M(a'),a), M(a'))]$ is identified via the \emph{edge g-formula} \citep{shpitser15hierarchy}:
{\small
\begin{equation}
\sum_{{\bf C},M,W} \mathbb{E}[Y | a', W,M,{\bf C}] p(W | a, M, {\bf C}) p(M | a', {\bf C}) p({\bf C}),
\label{eqn:edge-g-2}
\end{equation}
}
and may be estimated by plug-in estimators. PSEs such as (\ref{eqn:path-eff}) has been used in the context of observational studies of HIV patients for assessing the role of adherence in determining viral failure outcomes \citep{miles17quantifying}.

\section{Formalizing Discrimination And Prior Approaches To Fair Inference}
\label{sec:philosophy}

We are now ready to discuss prior approaches to fair inference.  In discussing the extent to which a particular approach is ``fair,''
we believe the gold standard is human intuition.  That is, we consider an approach inappropriate if it leads to counter-intuitive conclusions in examples.
For space reasons, we restrict attention to a representative subset of approaches.

A common class of approaches for fair inference is to quantify fairness via an associative (rather than causal) relationship between the sensitive feature $S$ and the outcome $Y$. 
For instance, \citep{feldman15certifying} adopted the \emph{80\% rule}, for comparing selection rates based on sensitive features.
This is a guideline (not a legal test) advocated by the Equal Employment Opportunity Commission \citep{EEOC} as a way of suggesting possible discrimination.
Rate of selection here is defined as the conditional probability of selection given the sensitive feature, or $p(Y | S)$.
\citep{feldman15certifying} proposed methods for removing disparities based on this rule via a link to classification accuracy.
A White House report on ``equal opportunity by design" \citep{WH} prompted \citep{hardt16equality} to propose a fairness criterion, called equalized odds, that ensures that true and false positive rates are equal across all groups.  This criterion is also associative. 

The issue with these approaches is they do not give intuitive results in cases where the sensitive feature is not randomly assigned (as gender is at conception), but instead exhibits spurious correlations with the outcome via another, possibly unobserved, feature.  We illustrate the difficulty with the following hypothetical example.  Certain states in the US prohibit discrimination based on past conviction history.  Prior convictions are influenced by other variables, such as gender (men have more prior convictions than women, on average). Consider a hypothetical dataset (consisting mostly of people with prior convictions) with two features -- prior conviction ($C$) and gender ($G$) as well as the hiring outcome ($H$). The values are coded as follows: male is $1$, female is $0$, prior conviction and hiring are $1$, lack of prior conviction and no hiring is $0$.  Assume the dataset is drawn from the joint density specified as follows: $p(G = 1) = 0.5$, and
{\small
\begin{align*}
\begin{tabular}{c|cc|c}
p(H=1| G,C) & G value & C value & p(C=1 |G) \\
\hline
0.06 & 1 & 1 & 0.99 \\
0.01 & 0 & 1 & 0.01 \\
0.2 & 1 & 0 &  \\
0.05 & 0 & 0 & \\
\end{tabular}
\end{align*}
}
That is, gender is randomly assigned at birth, the people in the cohort are very likely to have prior convictions (with men having more), and $p(H | C,G)$ specifies a certain hiring rule for the cohort.  For simplicity, we assume no other features of people in the cohort are relevant for either the prior conviction or the hiring decision. It's easy to show that
\[
p(H=1 | C=1) = 0.0595 \approx 0.0515 = p(H=1 | C=0).
\]
However, intuitively we would consider a hiring rule in this example fair if, in a hypothetical randomized trial that assigned convictions randomly (conviction to the case group, no conviction to the control group), the rule would yield equal hiring probabilities to cases and controls. In our example, this implies comparing counterfactual probabilities $p(H(C=1))$ and $p(H(C=0))$. 
Since we posited no other relevant features for assigning $C$ and $H$ than $A$, these probabilities are identified, via the adjustment formula described earlier, yielding $p(H(C=1)) = 0.035$, and $p(H(C=0)) = 0.125$.  That is, any method relying on associative measures of discrimination will likely conclude no discrimination here, yet the intuitively compelling test of discrimination will reveal a strong preference to hiring people without prior convictions.
The large difference between $p(H(C=0))$ and $p(H \mid C=0)$ has to do with extreme probabilities $p(C|G)$ in our example.
Even in less extreme examples, any approach that relies on associative measures of association will be led astray due to failing to properly model sources of confounding for the relationship of the sensitive feature and the outcome.
One might imagine that a simple repair in this example would be to also include $G$ as a feature.  The reason this does not work in general is not all features are possible to measure, and in general counterfactual probabilities are complex functions of the observed data, not just conditional densities \citep{shpitser06id}.

In the example above, it made intuitive sense to think of discrimination as a causal relationship between the sensitive feature and outcome.
In other examples, discrimination intuitively entails only a \emph{part} of the causal relationship.  Consider a modification of the hiring example above where potential discrimination is with respect to gender (a variable randomized at conception, which means worries about confounding are no longer relevant).  As before, consider binary variables $G$ and $H$ for gender and hiring, and an additional vector ${\bf C}$, representing applicant characteristics relevant for the job, of the kind that would appear on the resume.  The intuition here is it is legitimate to consider job characteristics in making hiring decisions \emph{even if} those characteristics are correlated with gender.  However, it is not legitimate to consider gender directly.  This intuition underscores resume ``name-swapping'' experiments where identical resumes are sent for review with names switched from a Caucasian sounding name to an African-American sounding name \citep{Bertrand2004}.  In such experiments, name serves as a proxy for race as a direct determinant of the hiring decision.

The definition of discrimination as related to causal pathways is further supported in the legal literature. The following definition of employment discrimination, which appeared in the legal literature \citep{carson96discrimination}, and was cited by \citep{pearl09causality}, makes clear the counterfactual nature of our intuitive conception of discrimination: 
\begin{quote}
The central question in any employment-discrimination case is whether the employer would have taken the same action had the employee been of a different race (age, sex, religion, national origin etc.) and everything else had been the same.
\end{quote}
The counterfactual ``had the employee been of a different gender'' phrase entails considering, for women, the outcome $Y$ had gender been male $G=1$, while
the ``everything else had been the same'' phrase entails considering job characteristics under the original gender $G=0$.  The resulting counterfactual
$Y(G=1,{\bf C}(G=0))$ is precisely the one used in mediation analysis to define natural direct effects.

It is possible to construct examples, discussed further, where some causal paths from a sensitive variable to the outcome are intuitively discriminatory, and others are not. Thus, our view is that discrimination ought to be formalized as the presence of certain path-specific effects.  The specific paths which correspond to discrimination are a \emph{domain specific issue}.  For example, physical fitness tests may be appropriate to administer for certain physically demanding jobs, such as construction, but not for white collar jobs, such as accounting.  As a result, a path from gender to the result of a test to a hiring decision may or may not be discriminatory, depending on the nature of the job.

Existing work considered similar proposals.  Prior work closest to ours appears in \citep{Zhang17causal}, where discrimination was also linked to path-specific effects.  While we agree with the link, we disagree on three essential points.  First, the authors do not appear to do any statistical inference and operate directly on discrete densities.  In high dimensional settings, where most practical outcome based inference takes place, statistical modeling becomes necessary, and an approach that avoids it will not scale.
In this paper we show how removing certain path-specific effects corresponds to a constrained inference problem on statistical models.  As we show later, the constrained optimization problems that arise are non-trivial.  Second, the authors propose an ad hoc repair in cases where the path-specific effect is not identifiable.  We believe this is a misunderstanding of the concept of non-identifiability.  If discrimination is indeed linked to a path-specific effect and this effect is not identifiable (not a function of the observed data), then the problem of removing discrimination is \emph{not solvable} without more assumptions.  In domains such as recidivism prediction, failure today and a better method tomorrow is preferable to an improper correction that preserves discriminatory practice.  We discuss more principled approaches to repairing lack of identifiability of discriminatory path-specific effects in later sections.  Finally, the authors, while repairing the observed data distribution to be fair, do not modify new instances to be classified in any way.  Since new instances are, by definition, drawn from the observed data distribution, which is ``unfair,'' no guarantees about discrimination when classifying new samples can be made.  We discuss this issue further below.

\section{Inference On Outcomes That Minimizes Discriminatory Path-Specific Effects}
\label{sec:inference}

We now describe our proposal precisely.  Assume we are interested in making inferences on outcomes given a joint distribution $p(Y,{\bf X})$ either modeled fully via a generative model, or partly via a discriminative model $p(Y,{\bf X}\setminus{\bf W}|{\bf W})$.
In addition, we assume
that $p(Y,{\bf X})$ is induced by a \emph{causal model} in the sense of \citep{pearl09causality,spirtes01causation},
that the presence of discrimination based on some sensitive feature $A$ with respect to $Y$ is represented by a PSE, and that this PSE is identified given the causal model as a functional $f(p(Y,{\bf X}))$.  Finally, we fix upper and lower bounds $\epsilon_l,\epsilon_u$ on the PSE, representing the degree of discrimination we are willing to tolerate.

Our proposal is to transform the inference problem on $p(Y,{\bf X})$ into an inference problem on another distribution $p^*(Y,{\bf X})$ which is close, in the Kullback-Leibler (KL)
divergence sense, to $p(Y,{\bf X})$ while also having the property that the PSE lies within $(\epsilon_l,\epsilon_u)$.  In some sense, $p^*$ represents a hypothetical ``fair world'' where discrimination is reduced, while $p$ represents our world, where discrimination is present.  A special case most relevant in practice is when $\epsilon_l = \epsilon_u$ is set to values that remove the PSE entirely.  We consider the more general case of bounding the PSE by $\epsilon_l,\epsilon_u$ to link with earlier work in \citep{Zhang17causal}, and for mathematical convenience.  In our framework, any function of $p$ of interest that we wish to make fair, such as the ACE or $\mathbb{E}[Y | {\bf X}]$, is to be computed from $p^*$ instead.  That is, just as causal inference is interested in hypothetical worlds representing randomized trials, so is fair inference interested in hypothetical worlds representing fair situations. And just as in causal inference, where it was important to only make inference in the hypothetical world of interest, it is important to make inferences only in the fair world, especially given new instances.  This is because new instances are likely drawn from the observed data distribution $p$, not from the hypothetical ``fair distribution'' $p^*$.  Since $p$ does not ensure discrimination is removed, any guarantees on discrimination removal made in $p^*$ will not translate to draws from $p$ for reasons similar to ones described in the covariate shift literature in machine learning -- the distribution of the new instance is not the right distribution.  We thus map any new instance ${\bf x}^i$ to a sensible version of it that is drawn from $p^*$.

A number of approaches for doing this mapping are possible.  In this paper we propose perhaps the simplest conservative approach.  That is, we will consider only generative
models for $p(Y,{\bf X}\setminus{\bf W}|{\bf W})$, and map to ``fair'' versions of this distribution $p^*(Y,{\bf X}\setminus{\bf W}|{\bf W})$.  This ensures that $p^*({\bf W})=p({\bf W})$,
and thus
${\bf x}^i_{\bf W}$ can be viewed as drawn from $p^*({\bf W})$, that is from the ``fair world.''
Since there is no unique way of specifying what values of ${\bf X}\setminus{\bf W}$ the ``fair version'' of the ${\bf x}^i$ instance would attain, we simply average over these possible values using $p^*$.  This amounts to predicting $Y$ using $\mathbb{E}^*[Y | {\bf W}]$ with ${\bf x}^i_{\bf W}$, and the expectation taken with respect to
$p^*(Y | {\bf W})$.

The choice of variables to include in ${\bf W}$, which governs the degree to which our model resembles a generative or a discriminative model is not obvious.  More discriminative models with larger ${\bf W}$ sets allow the use of larger parts of new instances for classification, which yields more information on the outcome.  On the other hand, more generative models with smaller ${\bf W}$ sets will be KL-closer to the true model.  To see this, consider two models, one which eliminates the discriminatory PSE by only constraining
$\mathbb{E}[Y | A,M,{\bf C}]$, and one which  eliminates the same PSE by constraining both $\mathbb{E}[Y | A,M,{\bf C}]$ and $p(M | A,{\bf C})$.  It is clear that the second model will be at least as KL-close to the true model $p(Y,{\bf X})$ as the first, and likely closer in general.  In this paper we propose a simple approach for choosing ${\bf W}$, based on the form of the estimator of the PSE, described further below.  We leave the investigation of more principled approaches for selecting ${\bf W}$ to future work.


A feature of our proposal is that we are selectively ignoring some known information about a new instance ${\bf x}^i$, if this information was drawn from the distribution that differs from $p^*$.  We believe this is unavoidable in fair inference settings -- the entire point is using the information ``as effectively as possible" is discriminatory.  We do want to use information as well as possible, but only insofar as we remain in the ``fair world".

\subsection{Fair Inference From Finite Samples}
\label{subsec:general}

Given a set of finite samples ${\cal D}$ drawn from $p(Y,{\bf X})$, a PSE representing discrimination identified as $f(p(Y,{\bf X}))$, a (possibly conditional) likelihood function ${\cal L}_{Y,{\bf X}}({\cal D}; {\bm\alpha})$ parameterized by $\bm\alpha$, an estimator $g({\cal D})$ of the PSE, and $\epsilon_l,\epsilon_u$, we approximate $p^*$ by solving a constrained maximum likelihood problem 
{\small
\begin{align}
\notag
\hat{\bm\alpha} = & \arg \max_{\bm\alpha} \hspace{0.2cm} {\cal L}_{Y, {\bf X}}({\cal D}; {\bm\alpha}) \\
 \text{subject to} & \hspace{0.2cm} \epsilon_l \leq g({\cal D}) \leq \epsilon_u.
 \label{eqn:c-mle}
\end{align}
}
Our choice for the set ${\bf W}$ will be guided by the form of the estimator $g(.)$.  Specifically ${\bf W}$ will contain variables with models not a part of $g(.)$. 
Since the estimators for the PSE developed within the causal inference literature do not model the baseline factors ${\bf C}$, ${\bf C} \subseteq {\bf W}$.  In addition,
certain estimators also do not use other parts of the model.  For example, (\ref{eqn:param-edge-g}) below does not use $p(A \mid {\bf C})$.  For such estimators we also
include those variables in ${\bf W}$.

%
%

We now illustrate the relationship between the choice of ${\bf W}$ and the choice of $g$ by considering three of the four consistent estimators of the NDE (assuming the model shown in Fig.~\ref{fig:med} (a) is correct) presented in \citep{tchetgen12semi2}.  The first estimator is the MLE plug in estimator for (\ref{eqn:edge-g}), given by
\begin{small}
\begin{align}
\frac{1}{n} \sum_{i,m} \left( \mathbb{E}[Y | a,m,{\bf c}^i
] - \mathbb{E}[Y | a',m,{\bf c}^i
] \right) 
p(m | a',{\bf c}^i
). 
\label{eqn:param-edge-g}
\end{align}
\end{small}
Since solving (\ref{eqn:c-mle}) using (\ref{eqn:param-edge-g}) entails constraining $\mathbb{E}[Y | A, M, {\bf C}]$ and $p(M | A, {\bf C})$, classifying a new point entails using
$\tilde{\mathbb{E}}[Y | A,{\bf C}
] = \sum_M \tilde{\mathbb{E}}[Y | A,M,{\bf C}
] \tilde{p}(M | A,{\bf C}
)$, where $\tilde{\mathbb{E}}$ and $\tilde{p}$ represent constrained models.

The second estimator uses all three models, as follows:
{\footnotesize
\begin{align}
\notag
\label{eqn:3-robust}
\frac{1}{n} \sum_i (
\frac{ a^i \ p(m^i \mid a',{\bf c}^i) \ \{ y_i - \mathbb{E}[Y | a,m^i,{\bf c}^i ] \}
}{ p(a|{\bf c}^i) \
	p(m^i | a,{\bf c}^i)}\\
\notag
+ \frac{(1 - a^i) \ \{ \mathbb{E}[Y | a,m^i,{\bf c}^i] - \eta(1,0,{\bf c}^i) \}
	}{p(a' | {\bf c}^i)}  + \eta(1,0,{\bf c}^i)\\
- \frac{(1-a^i)\ \{ y^i - \eta(0,0,{\bf c}^i) \} }{p(a' | {\bf c}^i 
)} +
	\eta(0,0,{\bf c}^i )), 
\end{align}
}
with 
$\eta(a,a',{\bf c}) \equiv \sum_m \mathbb{E}[Y | a,m,{\bf c}] \ p(m | a',{\bf c})$.
Since the models of $A,M$, and $Y$ are all constrained with this estimator, predicting $Y$ for a new instance is via $\tilde{\mathbb{E}}[Y | {\bf C}]$.
We discuss the advantages of this estimator in the next section.

The final estimator is based on inverse probability weighting (IPW). The IPW estimator uses the $A$ and $M$ models to estimate the NDE. We can fit the models
$p(A | \bf C
)$ and $p(M | A,\bf C
)$ by MLE, and use the following weighted empirical average as our estimate of the NDE:
{\small
\begin{align}
\frac{1}{n}  \sum_i
\left(
\frac{
a^i \ y^i \ p(m^i | A = 0, \mathbf{c}^i
)
}{
p(A = 1 | \mathbf{c}^i 
) \ p(m^i | A = 1,\mathbf{c}^i
)
}
- \frac{(1 - a^i) \ y^i}{p(A = 0 | \mathbf{c}^i
)}
\right).
\label{eqn-IPW}
\end{align}
}
Since solving the constrained MLE problem using this estimator entails only restricting parameters of $A$ and $M$ models, predicting a new instance $a^i,m^i,{\bf c}^i$ is done using
{\small
\begin{align}
\mathbb{E}[Y | {\bf C}
] = \sum_{A, M} \mathbb{E}[Y | A, M, {\bf C}
] \ p(M | A, {\bf C}
) 
\ p(A | {\bf C}
). 
\label{eqn:y-c}
\end{align}
}
A sensitive feature may affect the outcome through multiple paths, and paths other than a single edge path corresponding to the direct effect may be inadmissible.
Consider an example where $A$ is gender, and $Y$ is a hiring decision in a construction labor agency. We now consider two mediators of $A$, the number of children $M$, and physical strength as measured by an entrance test $W$.  In this setting, it seems that it is inappropriate for the applicant's gender $A$ to directly influence the hiring decision $Y$, nor for the number of the subject's children  to influence the hiring decision either since the consensus is that women should not be penalized in their career for the biological necessity of having to bear children in the family. However, gender also likely influences the subject's performance on the entrance test, and requiring that certain requirements of strength and fitness is reasonable in a job like construction.  The situation is represented by Fig.~\ref{fig:med} (b), with a hidden common cause of $M$ and $Y$ added since it does not influence the subsequent analysis.

In this case, the PSE that must be minimized for the purposes of making the hiring decision is given by (\ref{eqn:path-eff}), and is identified, given a causal model in Fig.~\ref{fig:med} (b), by (\ref{eqn:edge-g-2}).  If we use the analogue of (\ref{eqn:param-edge-g}), we would maximize
${\cal L}({\cal D}
; {\bm\alpha})$ subject to
{\small
\begin{align}
\notag
\frac{1}{n} \sum_{i=1}^n \sum_{w,m} &
\mathbb{E}[Y | 0,w^i,m^i,{\bf c}^i
] p(m^i | 0, {\bf c}^i
) \times \\
&
 \{ p(w^i | 1, m^i, {\bf c}^i
)-p(w^i | 0, m^i, {\bf c}^i
) \}
\label{eqn:path-mle}
\end{align}
}
being within $(\epsilon_l, \epsilon_u)$.  This would entail classifying new instances $a^i,w^i,m^i,{\bf c}^i$
using $\tilde{\mathbb{E}}[Y | A,{\bf C}
]$.  Our proposal can be generalized in this way to any setting where an identifiable PSE represents discrimination, using the complete theory of identification of PSEs \citep{shpitser13cogsci}, and plug-in MLE estimators that generalize (\ref{eqn:path-mle}).  We consider approaches for non-identifiable PSEs in one of the following sections.

\subsection{Fair Inference Via Box Constraints}
\label{subsec:box}

Generally, the optimization problem in (\ref{eqn:c-mle}) involves complex non-linear constraints on the parameter space.  However, in certain cases the optimization problem is significantly simpler, and involves box constraints on individual parameters of the likelihood.  In such cases, standard optimization software such as the \textit{optim} function in the R programming language can be used to directly solve (\ref{eqn:c-mle}).  We describe two such cases here.
First, if we assume a linear regression model for the outcome $Y = w_0 + w_a A + w_m M + w_c C$ in the causal graph of Fig.~\ref{fig:med}(a), then the
NDE on the mean difference scale, $\mathbb{E}[Y(1,M(0))] - \mathbb{E}[Y(0,M(0))]$ is equal to $w_a$, and (\ref{eqn:c-mle}) simplifies to a box constraint on $w_a$.  Note that setting $\epsilon_l = \epsilon_u = 0$ in this case coincides to simply dropping $A$ from the outcome regression.

Second, consider a setting with a binary outcome $Y$ specified with a logistic model, $\operatorname{logit}(P(Y = 1 \mid A, M, C)) = \theta_0 + \theta_aA + \theta_mM + \theta_cC $, a continuous mediator specified with a linear model, and the NDE (within a given level of $C$) defined on the odds ratio scale as in \citep{tyler10odds}:
{\small
\begin{align*}
\text{NDE} = \frac{ P(Y(1,M(0)) = 1)| C) / P(Y(1,M(0)) = 0)| C))}{P(Y(0,M(0)) = 1)| C) / P(Y(0,M(0)) = 0) | C))}. \nonumber 
\end{align*}
}

In this setting, it was shown in \citep{tyler10odds} that under certain additional assumptions, the NDE has no dependencies on $C$, and is approximately equal to $\exp(\theta_a)$. Hence, (\ref{eqn:c-mle}) can be expressed using box constraints on $\theta_a$.  Note that the null hypothesis of the absence of discrimination corresponds to the value of $1$ of the NDE on the odds ratio scale, and to $0$ on the mean difference scale.

\subsection{Fair Inference With A Regularized Outcome Model}
\label{subsec:complex-y}

In many applications, the goal of inference is not to approximate the true model itself, but to maximize out of sample prediction performance regardless of what the true model might be.  Validation datasets or resampling approaches can be used to assess the performance of such predictive models, with various regularization methods used to make the tradeoff between bias and variance.  The difficulty here is that searching for a model $Y$ with good out of sample prediction performance implies the true $Y$ model might be sufficiently complex that it may not lie within the model we consider.  This means we cannot use any estimator of the PSE that relies on the $Y$ model.  This is because most estimators that rely on the $Y$ model are not consistent if the $Y$ model is misspecified.  An inconsistent estimator of the PSE 
implies we cannot be sure solving the constrained optimization problem will indeed remove discrimination.

The key approach for addressing this is to use estimators that do not rely on the $Y$ model.  For the special case of the NDE, one such estimator is the IPW estimator, described earlier and in \citep{tchetgen12semi2}.  Another is the estimator in (\ref{eqn:3-robust}),
which was shown to be \emph{triply robust} in \citep{tchetgen12semi2}, meaning it remains consistent in the \emph{union model} where any two of the three models (of $Y$, $M$ and $A$) are specified correctly.  With either estimator, predicting a new instance $a^i,m^i,{\bf c}^i$ entails using $\tilde{\mathbb E}[Y | {\bf C}]$.  Since $A$ and $M$ models are assumed to be known, we regularize the model $\mathbb{E}[Y | A,M,{\bf C}]$ to maximize out of sample predictive performance using $\tilde{\mathbb{E}}[Y | {\bf C}]$.  Using these estimators ensures any regularization of the $Y$ model does not influence the estimate of the NDE, meaning that discrimination remains minimized. We give an example analysis illustrating this approach in the experiment section.

\subsection{Fairness In Computational Bayesian Methods}
\label{subsec:bayes}

Methods for fair inference described so far are fundamentally frequentist in character, in a sense that they assumed a particular true parameter value, and parameter fitting was constrained in a way that an estimate of this parameter was within specified bounds.  Here, we do not extend our approach to a fully Bayesian setting, where we would update distributions over causal parameters based on data, and use the resulting posterior distributions for constraining inferences.  Instead, we consider how Bayesian methods for estimating conditional densities can be adapted, as a computational tool, to our frequentist approach.

Many Bayesian methods do not compute a posterior distribution explicitly, but instead sample the posterior using Markov chain Monte Carlo approaches \citep{metropolis53equations}.  These sampling methods can  be used to compute any function of the posterior distribution, including conditional expectations, and can be modified to obey constraints in our problem in a straightforward way.
As an example, we consider BART, a popular Bayesian random forest method described in \citep{chipman10bart}.  This method constructs a distribution over a forest of regression trees, with a prior that favors small trees, and samples the posterior using a variant of Gibbs sampling, where a new tree is chosen while all others are held fixed.  A well known result \citep{gelfand92bayesian} states that a Gibbs sampler will generate samples from a constrained posterior directly if it rejects all draws that violate the constraint.

We implemented this simple method by modifying the R package (with a C++ backend) \textit{BayesTree}, and applied the result to the model in Fig \ref{fig:adultcompas} (a), where NDE was estimated via 
a mixed estimation strategy described in \citep{tchetgen12semi2}, where $A$ was assumed to be randomly assigned (i.e. no modeling, and hence no constraining, of $A$ was required), and the $Y$ model was fit using constrained BART. The experiment using the resulting constrained outcome model is described in the experimental section.

\section{Dealing With Non-Identification of the PSE}
\label{sec:non-id}

Suppose our problem entailed the causal model in Fig.~\ref{fig:med} (b), or Fig.~\ref{fig:med} (c) where in both cases only the NDE of $A$ on $Y$ is discriminatory.  Existing identification results for PSEs \citep{shpitser13cogsci} imply that the NDE is not identified in either model.  This means estimation of the NDE from observed data is not possible as the NDE is not a function of the observed data distribution in either model.

In such cases, three approaches are possible.  In both cases, the unobserved confounders $U$ are responsible for the lack of identification.  If it were possible to obtain data on these variables, or obtain reliable proxies for them, the NDE becomes identifiable in both cases.  If measuring $U$ is not possible, a second alternative is to consider a PSE that is identified, and that includes the paths in the PSE of interest and \emph{other paths}.  For example, in Fig.~\ref{fig:med} (b), while the NDE of $A$ on $Y$, which is the PSE including only the path $A \to Y$, is not identified, the PSE which includes paths $A \to Y$, $A \to M \to Y$, and $A \to M \to W \to Y$, namely (\ref{eqn:path-eff}), is.
The first counterfactual in the PSE contrast is identified in Fig.~\ref{fig:med} (b) by (\ref{eqn:edge-g-2}), and the second by the adjustment formula.

If we are using the PSE on the mean difference scale, the magnitude of the effect which includes more paths than the PSE we are interested in must be an upper bound on the magnitude of the PSE of interest in order for the bounds we impose to actually limit discrimination.  This is only possible if, for instance, all causal influence of $A$ on $Y$ along paths involved in the PSE are of the same sign.  In Fig.~\ref{fig:med} (b), this would mean assuming that if we expect the NDE of $A$ on $Y$ to be negative (due to discrimination), then it is also negative along the paths $A \to M \to W \to Y$, and $A \to M \to Y$.

If measuring $U$ is impossible, and it is not possible to find an identifiable PSE that includes the paths of interest from $A$ to $Y$, and serves as a useful upper bound to the PSE of interest, the other alternative is to use bounds derived for non-identifiable PSEs.  While finding such bounds is an open problem in general, they were derived in the context of the NDE with a discrete mediator in \citep{caleb16on}. 

The issue with non-identification of the PSE was also noted in \citep{Zhang17causal}. They proposed to change the causal model, specifically by cutting off some paths from the sensitive variable to the outcome such that the identification criterion in \citep{shpitser13cogsci} became satisfied, and the PSE became identified.
We  disagree with this approach, as we believe it amounts to ``redefining success.''  If the original causal model truly represents our beliefs about the structure of the problem, and in particular the pathways corresponding to discrimination, then making any sort of inferences in a model modified away from truth no longer tracks reality.  We would certainly not expect any kind of repair within a modified model to result in fair inferences in the real world.  The workarounds for non-identification we propose aim to stay within the true model, but try to obtain information on the true non-identified PSE, either by non-parametric bounds, or by including other pathways along with the ``unfair'' pathways.

\section{Experiments}
\label{sec:experiments}

We first illustrate our approach to fair inference via two datasets: the COMPAS dataset \citep{propublica} and the Adult dataset \citep{Lichman13adult}. We also
illustrate how a part of the model involving the outcome $Y$ may be regularized without compromising fair inferences if the NDE quantifying discrimination is estimated
using methods that are robust to misspecification of the $Y$ model.

\subsection{The COMPAS Dataset}
\label{sec:exp-COMPAS}

Correctional Offender Management Profiling for Alternative Sanctions, or COMPAS, is a risk assessment tool, created by the company Northpointe, that is being used across the US to determine whether to release or detain a defendant before his or her trial. Each pretrial defendant receives several COMPAS scores based on factors including but not limited to demographics, criminal history, family history, and social status. Among these scores, we are primarily interested in ``Risk of Recidivism". Propublica \citep{propublica} has obtained two years worth of COMPAS scores from the Broward County Sheriff's Office in Florida that contains scores for over $11000$ people who were assessed at the pretrial stage and scored in 2013 and 2014. COMPAS score for each defendant ranges from $1$ to $10$, with $10$ being the highest risk. 
Besides the COMPAS score, the data also includes records on defendant's age, gender, race, prior convictions, and whether or not recidivism occurred in a span of two years. We limited our attention to the cohort consisting of African-Americans and Caucasians. 

\begin{figure}[!t]
\centering
\begin{tikzpicture}[>=stealth, node distance=0.9cm]
    \tikzstyle{format} = [draw, very thick, circle, minimum size=5.0mm,
	inner sep=0pt]
	\begin{scope}
		\path[->, very thick]
			node[format] (A) {$A$}
			node[format, right of= A] (M) {$M$}
			node[format, right of= M] (Y) {$Y$}
			node[format, above of= M] (C) {$C$}
			(A) edge[blue] (M)
			(M) edge[blue] (Y)
			(A) edge[black!60!green, bend right] (Y)
			(C) edge[blue] (Y)
			(C) edge[blue] (M)
			(C) edge[blue] (A)
			node[below of=A, yshift=0.15cm, xshift=0.9cm] (l) {$(a)$} ;
	\end{scope}	
	\begin{scope}[xshift=3.4cm]
		\path[->, very thick]
			node[format] (A) {$A$}
			node[format, right of= A] (M) {$M$}
			node[format, right of= M] (L) {$L$}
			node[format, right of= L] (R) {$\bf R$}
			node[format, right of= R] (Y) {$Y$}
			node[format, above of= L] (C) {$\bf C$}
			node[format, gray, above of= M] (Umy) {$U_1$}
			node[format, gray, above of= R] (Ulr) {$U_2$}
			(A) edge[black!60!green] (M)
			(A) edge[blue, bend right=30] (L)
			(A) edge[blue, bend right=28] (R)
			(A) edge[black!60!green, bend left=20] (Y)
			(M) edge[black!60!green] (L)
			(M) edge[black!60!green, bend left] (R)
			(M) edge[black!60!green, bend right=28] (Y)
			(L) edge[black!60!green] (R)
			(L) edge[black!60!green, bend left=50] (Y)
			(R) edge[black!60!green] (Y)
			(C) edge[blue] (M)
			(C) edge[blue] (L)
			(C) edge[blue] (R)
			(C) edge[blue, bend left=7] (Y)
			(Umy) edge[red] (M)
			(Umy) edge[red, bend left=60] (Y)
			(Ulr) edge[red, bend right=20] (L)
			(Ulr) edge[red] (R)
			node[below of=L, yshift=0.15cm, xshift=0.0cm] (l) {$(b)$}
			;
	\end{scope}
\end{tikzpicture} 
\caption{Causal graphs for (a) the COMPAS dataset, and (b) the Adult dataset.}
\label{fig:adultcompas}
\end{figure}
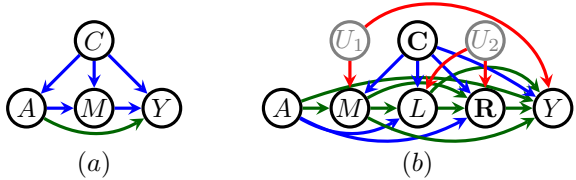

We are interested in predicting whether a defendant would reoffend using the COMPAS data.  For illustration, we assume the use of prior convictions, possibly influenced by race, is fair for determining recidivism.  Thus, we defined discrimination as effect along the direct path from race to the recidivism prediction outcome. The simplified causal graph model for this task is given in Figure \ref{fig:adultcompas} (a), where $A$ denotes race, prior convictions is the mediator $M$, demographic information such as age and gender are collected in $\bf C$, and $Y$ is recidivism.   The ``disallowed" path in this problem is drawn in green in Figure \ref{fig:adultcompas}(a).  The effect along this path is the NDE.  The objective is to learn a fair model for $Y$.  i.e. a model where NDE is minimized. 

We obtained the posterior sample representation of $\mathbb{E}[Y | A,M,{\bf C}]$ via both regular and constrained BART.
Under the unconstrained posterior, the NDE (on the odds ratio scale) was equal to $1.3$.  This number is interpreted to mean that
the odds of recidivism would have been $1.3$ times higher had we
changed race from Caucasian to African-American. 
In our experiment we restricted NDE to lie between $0.95$ and $1.05$. Using unconstrained BART, our prediction accuracy on the test set was $67.8 \%$,  removing treatment from the outcome model dropped the accuracy to $64.0\%$, and using constrained BART lead to the accuracy of $66.4\%$. As expected, dropping race, an informative feature, led to a greater decrease in accuracy, compared to simply constraining the outcome model to obey the constraint on the NDE.

In addition to our approach to removing discrimination, we are also interested in assessing the extent to which the existing recidivism classifier used by Northpointe is biased. Unfortunately, we do not have access to the exact model which generated COMPAS scores, since it is proprietary, nor all the input features used. Instead, we used our dataset to predict a binarized COMPAS score by fitting the model $\tilde{p}(Y | M,{\bf C})$ using BART.  We dropped race, as we know Northpointe's model does not use that feature. Discrimination, as we defined it, may still be present even if we drop race.  To assess discrimination, we estimate the NDE, our measure of discrimination, in the semiparametric model of $p(Y,M,A,{\bf C})$, where the only constraint is that $p(Y | M,{\bf C})$ is equal to $\tilde{p}$ above.  This model corresponds to (our approximation of) the ``world'' used by Northpointe.  Measuring the NDE on the ratio scale using this model yielded $2.1$, which is far from $1$ (the null effect value). In other words, assuming the defendant is Caucasian, then the odds of recidivism for him would be $2.1$ times higher had he been, contrary to fact, African-American. Thus, our best guess on Northpointe's model is that it is severely discriminatory. 
 
\subsection{The Adult Dataset}
The ``adult'' dataset from the UCI repository has records on $14$ attributes such as demographic information, level of education, and job related variables such as occupation and work class on $48842$ instances along with their income that is recorded as a binary variable denoting whether individuals have income above or below $50k$ -- high vs low income. The objective is to learn a statistical model that predicts the class of income for a given individual. Suppose  banks are interested in using this model to identify reliable candidates for loan application. Raw use of data might construct models that are biased towards females who are perceived to have lower income in general compared to males. The causal model for this dataset is drawn in Figure \ref{fig:adultcompas}(b). Gender is the sensitive variable in this example denoted by $A$ in figure \ref{fig:adultcompas}(b) and income class is denoted by $Y$. $M$ denotes the marital status, $L$ denotes the level of education, and $\bf R$ consists of three variables, occupation, hours per week, and work class. The baseline variables including age and nationality are collected in $\bf C$. $U_1$ and $U_2$ capture the unobserved confounders between $M,Y$ and $L, \bf R$, respectively. 

Here, besides the direct effect ($A\rightarrow Y$), we would like to remove the effect of gender on income through marital status ($A \rightarrow M \rightarrow \hdots \rightarrow Y$). The ``disallowed" paths are drawn in green in Figure \ref{fig:adultcompas}(b). 
The PSE along the green paths is identifiable via the recanting district criterion in \citep{shpitser13cogsci}, and can be computed by calculating odds ratio or contrast comparison of the counterfactual variable
$Y( a,  M(a),  L(a', M(a) ),  {\bf R} (a', M(a),  L(a', M(a) ) ), {\bf C}),$ where $a'$ is set to a baseline value, $a=1$ in one counterfactual, and $a=0$ in the other. The  counterfactual distribution can be estimated from the following functional: 
$\sum_{{\bf V} \setminus A} \{  p(Y | a, m, l, {\bf r, c}) \prod_{i = 1}^{3} p(r_i | a', m, l, {\bf c}) p(l | a', m, {\bf c})$
$p(m | a, {\bf c}) \ p(\bf c) \}$,
where ${\bf V}$ are all observed variables.

If we use logistic regression to model $Y$ and linear regression to model other variables given their past, and compute the PSE on the odds ratio scale, it is straightforward to show that the PSE simplifies to
$\exp\big( \T_a^y + \T_m^y\T_a^m + \T_l^y\T_m^l\T_a^m+ \sum_i\T_{r_i}^y(\T_m^{r_i}\T_a^m + \T_l^{r_i}\T_m^l\T_a^m)\big)$, where $\T_i^j$ denotes the coefficient associated with variable $i$ in modeling the variable $j$, \citep{tyler10odds}. Therefore, the constraint in (\ref{eqn:c-mle}) is an easy function to compute, and the resulting constrained optimization problem relatively easy to solve.

We trained two models for $Y$, one by maximizing the constrained likelihood in (\ref{eqn:c-mle}) using the R package \textit{nloptr}, and the other by using the full model with no constrain. For performance evaluation on test set, we should use $\mathbb{E}[Y | A,{\bf C}]$ in constrained model and $\mathbb{E}[Y | A, M, L, {\bf R, C}]$ in unconstrained model. 

The PSE in the unconstrained model is $3.16$. This means, the odds of having a high income would have been more than 3 times higher for a female if her sex and marital status would have been the same as if she was a male. We solve the constrained problem by restricting the PSE, as estimated by (\ref{eqn:path-mle}), to lie between $0.95$ and $1.05$. Accuracy in the unconstrained model is $82\%$, and drops to $72\%$ in the constrained model while assuring that the constrained model is fair. 

A very simple and naive approach to the problem that, superficially, might appear to be sensible for removing discrimination, is to drop the sensitive feature, which in this case results in the same accuracy as the unconstrained model. However, we believe dropping the sensitive feature from the model is a poor choice in our setting, both because sensitive features are often highly predictive of the outcome in interesting (and politicized) cases, and because dropping the sensitive feature does not in fact remove discrimination (as we defined it)!

\subsection{Selecting The Outcome Model To Maximize Out Of Sample Predictive Performance}

The search for an outcome model with the best out of sample performance, as is often done in machine learning problems, may result in a model which does not give consistent estimates of the NDE and thus does not guarantee removal of discrimination, if the NDE estimator is not chosen carefully.
As discussed in the previous sections, the key approach is to use estimators that do not rely on the $Y$ model being correctly specified, such as the triply robust estimator, and the IPW estimator.  Here we demonstrate, via a simple simulation study, that selecting the outcome model to maximize predictive performance does not interfere with solving the constrained optimization problem for removing discrimination as long as we use triply robust or IPW estimators given that $A$ and $M$ models are specified correctly. 

 We generated $4000$ data points using the models shown in (\ref{eqn:sim1}) and split the data into training and validation sets.  
{\small
\begin{align}
A &\sim \text{Bernoulli}(p = 0.5) \nonumber \\
C_1, C_2 &\sim \mathcal{N}\left(\mu = (0, 0), \Sigma = 
\begin{pmatrix}
  2 & 1 \\ 1 & 2  
 \end{pmatrix} \right)  \nonumber \\
\text{logit}(p(M)) &\sim -3 + 0.8C_1 + 0.7C_2 + 0.3A + 0.3AC_1 + \nonumber \\&
 - 0.3AC_2  \nonumber \\
Y = 5 + 3A + & C_1 + 0.3C_2 + 0.8M +  0.5A(C_1 + C_2 + M)  \nonumber \\
+ 0.4C_1C_2  &+ 0.2M(C_1 + C_2) + \mathcal{N}(0, 2). 
\label{eqn:sim1}
\end{align}
} 

We assume $A$ and $M$ models are correctly specified; $A$ is randomized (like race or gender) and $M$ has a logistic regression model with interaction terms. Using the IPW estimator, which only uses $A$ and $M$ models, we obtain the NDE (on the ratio scale) of $3.01$.  As expected, the triply robust estimator, which uses $A, M$, and $Y$ models,
gives us the same estimate of NDE,  even under a misspecified $Y$ model. 

To select the most predictive outcome model, we searched over possible candidate models and performed constrained optimization that restricted the NDE to be within $-0.5$ and $0.5$. We chose the model that lead to the smallest rMSE on the test data, where the predictions were done using
$\tilde{\mathbb{E}}[Y | {\bf C}]$, with this expectation evaluated by marginalizing over the candidate outcome model $\mathbb{E}[Y | A,M,{\bf C}]$, and the constrained models for
$M$ and $A$.  Our pool of candidate models were the linear regression models with different subsets of interaction terms. 
As expected, when using the triply robust estimator under the correctly specified models for $A$ and $M$,
the NDE for all candidate $Y$ models we considered was almost the same and close to the truth ($3.01$).
Out of the pool of candidate models, the following model was selected to have the smallest rMSE:
{\small
\begin{align}
\mathbb{E}[Y \mid {\bf D}] 
= 5.1 + 2.86 A + 1.2 C_1 + 0.5 C_2 +  0.39 C_1 C_2 + \nonumber \\
 0.15 M + 1.74 AM + 0.29 MC_1 + 0.02 M C_2,   \nonumber 
\end{align}
}
where ${\bf D} \subseteq {\bf C}$.
Consider, by contrast, what happened when we used an estimator that relied on the $Y$ model being specified correctly.
We pick the following (incorrect) $Y$ model:
{\small
\begin{align}
\mathbb{E}[Y \mid {\bf D}] 
= \theta_0 + \theta_a A + \sum_{i}\theta_{i} C_i  + \theta_m M + \theta_{ac} (AC_1)^2, \nonumber 
\end{align}
}
and compute the NDE using (\ref{eqn:edge-g}), to obtain the value of $2.7$. 

Performing the constrained optimization using the above model and the estimator in (\ref{eqn:edge-g}) would lead us to the optimal coefficients for the $M$ and $Y$ models that ensure the NDE is within $(-0.5,0.5)$, as desired.  However, since the $Y$ model was incorrect, and (\ref{eqn:edge-g}) was not robust to misspecification of $Y$, the results cannot be trusted.  Indeed, using the constrained coefficients for $M$ and $Y$ models in the triply robust estimator, that is robust to misspecified of $Y$, leads to a large NDE of $3.07$.

The takeaway here is that the classical machine learning task of model selection to optimize out of sample prediction performance, be it via parameter regularization or other methods, can only ensure fairness if the estimators for the degree of fairness, as quantified by the PSE, do not rely on the model being selected, the models the estimators do rely on are specified correctly, and only the part of the model the estimators do not rely on is selected.

\section{Discussion And Conclusions}
\label{sec:discussion}
In this paper, we considered the problem of fair statistical inference on outcomes, a setting where we wish to minimize discrimination with respect to a particular sensitive feature, such as race or gender.  We formalized the presence of discrimination as the presence of a certain \emph{path-specific effect (PSE)} \citep{pearl01direct,shpitser13cogsci}, as defined in mediation analysis, and framed the problem as one where we maximize the likelihood subject to constraints that restrict the magnitude of the PSE.  We explored the implications of this view for predicting outcomes out of sample, for cases where the PSE of interest is not identified, and for computational Bayesian methods. We illustrated our approach using experiments on real datasets. 

One of the advantages of our approach is it can be readily extended to concepts like affirmative action and ``the wage gap'' in a way that matches human intuition. To conceptualize affirmative action, we propose to define a set of ``valid paths" from $A$ (race/sexual orientation) to $Y$ (admission decision), perhaps paths through academic merit, or extracurriculars, or even the direct path, and solve a constrained optimization problem that \emph{increases} the PSE along these paths. Here we mean placing a lower bound $\epsilon_l$ on the PSE away from the value corresponding to ``no effect".  Then, we learn $p^*$ as the KL-closest distribution to the observed data distribution $p$ that satisfies the constraint on the PSE.  Finally, we predict the admission decision of a new instance $\bf X$ in a similar way as the proposal in our paper, by using the information in the new instance $\bf X$ shared between $p$ and $p^*$, and predicting/averaging over other information using $p^*$. We thus ``count the causal influence of the sensitive feature on admission via prescribed paths" more highly among disadvantaged minorities.  Defining these paths is a domain-specific issue. Increasing the PSE potentially lowers predictive performance, just as decreasing the PSE did in our experiments on reducing discrimination.  This makes sense since we are moving away from the PSE implied by the ``unfair world" given by the MLE towards something else that we deem more ``fair".  A similar definition can be made for ``the wage gap", which we believe should be meaningfully defined as a comparison of the PSE of gender on salary with respect to ``inappropriate paths.''

One methodological difficulty with our approach is the need for a computationally challenging constrained optimization problem.
An alternative would be to reparameterize the observed data likelihood to include the causal parameter corresponding to the discrimination PSE, in a way causal parameters have been added to the likelihood in structural nested mean models \citep{robins99marginal}.  Under such a reparameterization, minimizing the PSE always corresponds to imposing box constraints on the likelihood.  However, this reparameterization is currently an open problem. 

\section{Acknowledgments}
The research was supported by the grants R01 AI104459-01A1 and R01 AI127271-01A1.  We thank James M. Robins, David Sontag and his group, Alexandra Chouldechova, and Shira Mitchell for insightful conversations on fairness issues.  We also thank the anonymous reviewers for their comments that greatly improved the manuscript. 

{\small
\bibliography{references}
\bibliographystyle{aaai}
}

\end{document}